\documentclass[runningheads]{llncs}
\usepackage[T1]{fontenc}
\usepackage{graphicx}
\usepackage{booktabs}
\usepackage[misc]{ifsym}
\newcommand{\corr}{(\Letter)}
\usepackage{amsmath, amssymb}
\usepackage{hyperref}
\usepackage{booktabs,multirow,wrapfig,lipsum, tabularx}
\usepackage{float}
\usepackage{enumitem}
\usepackage{xfrac}
\usepackage[font=small,skip=10pt]{caption}

\begin{document}

\title{From Lab to Factory: Pitfalls and Guidelines for Self-/Unsupervised Defect Detection on Low-Quality Industrial Images}
% SH: Title will most likely change a bit, too.

\titlerunning{Self-/Unsupervised Low-Quality Image Defect Detection}
% If the full title of your paper is short enough to also fit in the running head, you can omit the abbreviated paper title here. You can check as follows: if you comment out the \titlerunning line, something will appear in the header of all odd-numbered pages of your PDF from page 3 onward. This something is either the full title (in which case all is well), or the error message "Title Suppressed Due to Excessive Length". If this error message appears, you're going to want to provide an abbreviated title within the \titlerunning command, because if you won't do it, Springer will do it for you.

%N.B.: Author information (both in the \author{} and \authorrunning{} command) should only be present in the Camera-Ready Version of your paper. The version that you initially submit for review, ought to be double-blind. So, when initially submitting your paper, use:
%\author{Author information scrubbed for double-blind reviewing}
\author{Sebastian Hönel\inst{1}\corr\orcidID{0000-0001-7937-1645}
\and Jonas Nordqvist\inst{1}\orcidID{0000-0002-0510-6782}}
\tocauthor{Sebastian Hönel, Jonas Nordqvist}
\toctitle{From Lab to Factory: Pitfalls and Guidelines for Self-/Unsupervised Defect Detection on Low-Quality Industrial Images}
% You may leave out the orcidID information, if you want to.
% Use \corr to indicate the corresponding author. Note the spacing around the \corr command. Only one author can be the corresponding author.

%N.B.: comment out the \authorrunning{} command for the double-blind version of your paper submitted for review. Later, if your paper is accepted, use the command for the Camera-Ready Version.
\authorrunning{S. Hönel and J. Nordqvist}
% If there are two authors, write 'A.L. Benjamin and C.C. Broadus Jr.'
% If there are more than two authors, '[...] et al.' is used.

\institute{Linn\ae us University, Växjö, Sweden \email{\{sebastian.honel,jonas.nordqvist\}@lnu.se}}
%\and
%Fictional West Coast University, Long Beach CA 90840, USA \email{ccb@fwcu.fake}
%\and
%Kwik-E-Mart, Germany
%\email{lncs@springer.com}}

\maketitle              % typeset the header of the contribution

\begin{abstract}
% SH: The abstract will VERY LIKELY change, I just copy-paste it here.
%The abstract should briefly summarize the contents of the paper in 150--250 words.
The detection and localization of quality-related problems in industrially mass-produced products has historically relied on manual inspection, which is costly and error-prone. Machine learning has the potential to replace manual handling. As such, the desire is to facilitate an unsupervised (or self-supervised) approach, as it is often impossible to specify all conceivable defects ahead of time. A plethora of prior works have demonstrated the aptitude of common reconstruction-, embedding-, and synthesis-based methods in laboratory settings. However, in practice, we observe that most methods do not handle low data quality well or exude low robustness in unfavorable, but typical real-world settings.
For practitioners it may be very difficult to identify the actual underlying problem when such methods underperform.
Worse, often-reported metrics (e.g., AUROC) are rarely suitable in practice and may give misleading results.
In our setting, we attempt to identify subtle anomalies on the surface of blasted forged metal parts, using rather low-quality RGB imagery only, which is a common industrial setting.
We specifically evaluate two types of state-of-the-art models that allow us to identify and improve quality issues in production data, without having to obtain new data.
Our contribution is to provide guardrails for practitioners that allow them to identify problems related to, e.g., (lack of) robustness or invariance, in either the chosen model or the data reliably in similar scenarios. Furthermore, we exemplify common pitfalls in and shortcomings of likelihood-based approaches and outline a framework for proper empirical risk estimation that is more suitable for real-world scenarios.

%\keywords{First keyword  \and Second keyword \and Another keyword.}
\end{abstract}

% SH: @Jonas: I enabled the TOC while we write, so we have a good overview. Will remove it later!
%\newpage % SH: TODO: <- REMOVE!
%\setcounter{tocdepth}{5} % SH: TODO: <- REMOVE!
%\tableofcontents % SH: TODO: <- REMOVE!

\section{Introduction}
% SH: Some notes:
% Keys:
% -	Unsupervised anomaly detection
% -	Subtle anomalies
% -	Low data quality, lack of robustness
%   o	Problematic camera and backgrounds
%   o	Many different angles
% -	Treacherous methods and deceiving reporting
%   o	Husky-Snow problem: NFs assign probabilities > 0 to black pixels
%   o	AUC is problematic
% -	Data preparation; segmentation
%   o	Invariance and equivariance on image or object level
% -	Benchmark methods that are unstable in practice (e.g., fragile NFs)
%   o	Also: Invariance (or lack of), equivariance in methods
% -	Drawing a not-so-random sample (NFs cannot draw outliers)
%   o	Density estimation not per se suitable for AD
%
%
% SH: Setting the stage; what is AD/AL and what is the typical industrial scenario.
Industrial manufacturing requires quality assurance. A common approach to quality evaluation is through manual visual inspection, which is expensive, repetitive, and error-prone~\cite{GuRD24}.
Recent advances in deep learning and computer vision have boosted interest in automated self- or unsupervised anomaly detection methods~\cite{Chai2021_CV}.
In such approaches, acquiring nominal (defect-free) data is affordable, but exhaustively defining abnormal defect variants often proves difficult or impossible.
Thus, anomaly detection (AD) is framed as an out-of-distribution challenge, distinguishing nominal samples from those outside the known distribution~\cite{RothPZSBG22_TowardsTotal}.
A related task, anomaly localization (AL), further identifies where anomalies occur, benefiting interpretability by providing visual cues and enabling human-in-the-loop systems~\cite{TaoGZYA22_AnomLocal}.

AD/AL methods in visual inspection generally fall into three categories: embedding-, reconstruction-, and synthesis-based approaches~\cite{ChenLLZ24_GLASS}.
Embedding-based approaches use pretrained feature extractors. % (e.g., CNNs or vision transformers~\cite{RudolphWRW22_CSFLOW,AlberHB24_VitTransf}) to convert raw data into suitable representations.
Reconstruction approaches exploit differences between input and reconstructed images, or encodings and decodings thereof, to detect anomalies~\cite{WangLLLZ24_POUTA}.
Synthesis methods train classifiers by generating artificial defects to differentiate them from nominal samples, effectively combining unsupervised and semi-supervised approaches~\cite{ZavrtanikKS21_DRAEM}.
Semi-supervised approaches are now generally preferred.
Unsupervised methods often lack reliability in precisely detecting out-of-distribution cases, whereas fully-supervised methods require costly data labeling, suffer limited data availability, and cannot easily handle unknown defect types or label noise~\cite{TaoGZYA22_AnomLocal}.

Visual-inspection AD methods are commonly evaluated on standard datasets such as MVTec AD~\cite{BergmannBFSS21_MVTEC_AD} or Magnetic Tile Defects (MTD)~\cite{HuangQY20_MTD}. %, and Kolektor SDD~\cite{Tabernik2019_Kolektor_SDD}.
However, these benchmark datasets largely come from controlled laboratory environments.
Hence, most images feature simple setups, single centered objects with uniform backgrounds, static orientations and camera distances, and lack realistic disturbances such as reflections, shadows, or blur.
Recently, Jezek et al.~\cite{JezekJBDS21_MPDD} introduced the Metal Parts Defect Detection (MPDD) dataset, a more challenging real-world scenario encompassing some typical practical issues.
Their evaluation demonstrates a significant deterioration of state-of-the-art performance, especially for image-level AD, highlighting the gap between lab-based benchmarks and real production environments.

In view of this gap, this paper explores practical problems related to data quality, modeling choices, and proper evaluation.
Important data-related challenges include excessively subtle anomalies relative to natural variance, unwanted background disturbances, reflections, and other unfavorable image-capturing conditions.
Additionally, certain conditions can impact models differently.
While introducing rotations or varying viewpoints can broaden data diversity and generalization, the effectiveness of these augmentations strongly depends on the model chosen.
Some models handle image variations like object scale or position robustly, while others decline significantly in performance if these transformations occur among the objects themselves.

Regarding performance evaluation, anomaly detection results frequently use area under the receiver operating characteristic curve (AUROC).
Although AUROC is valuable when comparing methods within identical datasets, interpreting reported AUROC values across different publications or scenarios should be done with caution.
Its interpretation strongly depends on the underlying data and evaluation strategies. Often only maximum or average AUROC scores are reported, lacking uncompromising replication.
Importantly, classifiers built from selected thresholds for anomaly scores are rarely evaluated rigorously using proper repeated cross-validation or bootstrapping methods that provide reliable estimates of empirical risk and realistic performance in deployment scenarios. 

\vspace{5pt}
\noindent
Our contributions are summarized as follows:
\begin{enumerate}
    \item We overview critical data-acquisition and data-quality issues encountered in practice, and suggest directions for improvement.
    \item We investigate how core architectural components in current AD/AL methods respond differently to specific real-world issues, offering practical recommendations for properly pairing datasets with models.
    \item We propose an evaluation framework based on adequate outer resampling to accurately estimate empirical model risk and derive confidence intervals, better predicting factory-floor model performance. 
\end{enumerate}

\noindent
The remainder of this paper is structured as follows.
In Section~\ref{sec:bg-and-related}, we present the relevant background, related work, and describe the concrete dataset and problem we attempt solving.
In Section~\ref{sec:limits-of-nfs}, we elucidate limitations of flow-based architectures.
Section~\ref{sec:robust-estimation} is dedicated to outlining and validating a robust estimation framework.
Lastly, Section~\ref{sec:discuss-fw} offers a discussion and an outlook on future work.

\section{Background and Related Work}\label{sec:bg-and-related}
Anomaly detection in imagery is a critical task in various domains, including industrial inspection, medical imaging, and security, where identifying irregularities is essential. Due to the rarity and diversity of anomalies, unsupervised and semi-supervised learning approaches are commonly employed, as they enable detection without the need for extensively labeled (complementary) datasets.

Unsupervised methods typically identify anomalies by modeling the distribution of normal samples. One widely used approach involves reconstruction-based techniques, such as autoencoders~\cite{Hinton2006}, where the model is trained to reconstruct normal data, and anomalies are inferred from high reconstruction errors~\cite{Sakurada2014}. Another method leverages generative adversarial networks (GANs)~\cite{Goodfellow2014}, assessing deviations from the learned distribution to detect anomalies~\cite{Park2023}.
A third strategy involves normalizing flows, which estimate the likelihood of feature maps and classify low-likelihood regions as anomalous~\cite{RudWan2021,RudolphWRW22_CSFLOW}, where the latter case is further discussed in Section~\ref{sec:csflow}. Additionally, clustering techniques such as $k$-means and DBSCAN can be utilized, where data points identified as outliers are considered anomalies.

In the context of self-supervised learning, techniques such as synthetic anomaly generation or auxiliary tasks have been explored to enhance feature representation learning, thereby improving anomaly detection performance (e.g.,~\cite{Hendrycks2019}). % golan2018
The current state-of-the-art in the realm of AD is a method which utilizes several techniques in AD. This method is further described in Section~\ref{sec:glass}.

% Anomaly detection in general !!!!!

% SH: UPDATE: 
% CS-Flow has its own section already, and I am adding one for GLASS directly afterwards (currently 4.1 and 4.2).
% Here, however, we would need a small but broad overview of unsupervised/semi-supervised AD methods.
% Here, we should just especially mention CS-Flow and GLASS and include a forward reference to those respective subsections.

\subsection{Normalizing Flows}
Let~$ d \geq 1 $ be an integer, and consider a vector~$ \mathbf{x} \in \mathbb{R}^d $. A normalizing flow models the distribution of~$ \mathbf{x} $ using a diffeomorphic transformation~$ T $, which is bijective and smooth along with its inverse. The transformation is applied to a latent variable~$ \mathbf{z} $ drawn from a base distribution~$ p_{\mathbf{z}}(\mathbf{z}) $. Given~$ \mathbf{x} = T(\mathbf{z}) $, the density of~$ \mathbf{x} $ follows from the change-of-variables formula
\[
p_\mathbf{x}(\mathbf{x}) = p_{\mathbf{z}}(\mathbf{z}) \left| \det J_T(\mathbf{z}) \right|^{-1}, \quad \text{where } \mathbf{z} = T^{-1}(\mathbf{x})
\]
and~$J_T(\mathbf{z})$ is the Jacobian of~$ T $. In practice,~$ T $ is parameterized, and
%, often using a neural network.
in the context of normalizing flows, the set of parameters $\theta$ for a transformation $T$ is learned by a so-called \emph{conditioner}, typically implemented by some fully-connected neural network.
Training a normalizing flow involves minimizing the forward Kullback--Leibler divergence between the data distribution~$ p_\mathbf{x}(\mathbf{x}) $ and the transformed base distribution~$ \widehat{p}_\mathbf{x}(\mathbf{x}; \theta) $. The loss function is:
\begin{align*}\label{eqn:loss}
\mathcal{L}(\theta) &= D_{\text{KL}}\left(p_\mathbf{x}(\mathbf{x}) \, \big\| \, \widehat{p}_\mathbf{x}(\mathbf{x}; \theta)\right) = -\mathbb{E}_{p_\mathbf{x}(\mathbf{x})}\left[\log\left(\widehat{p}_\mathbf{x}(\mathbf{x}; \theta)\right)\right] + C\\
&= -\mathbb{E}_{p_\mathbf{x}(\mathbf{x})}\left[\log\left(p_\mathbf{z}\left(T^{-1}(\mathbf{x}; \theta)\right)\right) 
+ \log\left|\det J_{T^{-1}}(\mathbf{x})\right|\right] + C ,
\end{align*}
where~$ C $ is a constant. This may be estimated via Monte Carlo sampling. %, and minimizing it corresponds to maximizing the likelihood of~$\theta$. 
For further details, see, e.g.,~\cite{papamakarios2021}. % Kobyzev2020

\subsection{Some Problems of Purely Likelihood-Based Approaches of AD}
Likelihood-based deep generative models have been widely used in anomaly detection. However, there are a number of critical limitations which makes these prone to failure.
One of these problems is that the model may assign higher likelihood to data which may be semantically different although still normal in some sense.
For instance, models trained on Cifar-10 have been shown to assign higher likelihoods to images from SVHN.
One explanation to this problem has been proposed in \cite{CateriniL21}, where the argument is that there is a dependence on the entropy of the dataset, which gives rise to unwanted behavior.
%
%For instance, models trained on CIFAR-10\footnote{A dataset of 60,000 small ($32$x$32$) color images across 10 object classes, commonly used for image classification tasks.} have been shown to assign higher likelihood to images in SVHN\footnote{A dataset of real-world digit images ($32$x$32$) extracted from house number plates, used for digit recognition in machine learning.} (OOD) than in-distribution images in CIFAR-10 \cite{NalisnickMTGL19}. 
%One explanation to this problem has been proposed in \cite{CateriniL21}, where the argument is that there is a dependence on the entropy of the dataset, which gives rise to unwanted behavior.
%%% kanske ta bort fotnoterna här!!
%
Another problem is that likelihood-based models struggle in high-dimensional spaces where background statistics dominate the likelihood estimation. However, to circumvent these issues it is suggested in \cite{Ren2019} to use likelihood ratios instead of crude likelihoods to handle the background statistics.

\subsection{Our Problem and Dataset}\label{sec:prob-and-dataset}
% SH: @Jonas: If you could fill out as many of those blanks as possible, it'd be nice :)
%
% Describe our industrial cast-iron part dataset clearly & succinctly.
% - Object description (coupling links for chains)
% - Camera setup/conditions (flash, reflections, background metal plate, rotation angles)
% - Demonstrate visually (include representative images), highlighting variable views, backgrounds, and subtle defects.

\begin{figure}
    \center
    \includegraphics[width=\textwidth]{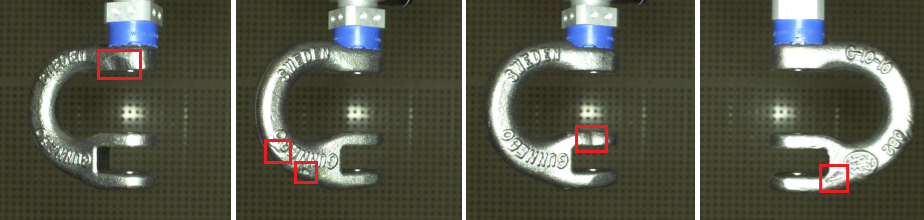}
    \caption{\label{fig:dataset-example} Four defective examples from the G-link dataset. Anomalies are coarsely marked using a red bounding box (zoom for more details).}
\end{figure}

The dataset consists of medium-resolution images of blasted forged steel coupling links (``G-links'') captured using an area-scan camera as the coupling link was translated and rotated by an industrial robot~\cite{g_link_dataset}.
This means that each individual piece is captured by roughly 120 images over different angles.
Damaged pieces were collected using the regular manual inspection process in the production environment over an extended time frame.
In the training set, there are $3,424$ images of non-defect coupling links for training.
The validation set is comprised of $568$ images, $165$ of which contain a defect.
Figure~\ref{fig:dataset-example} shows four defective samples with surface dents, deeper deformations, and scratches.
These kind of anomalies arise naturally in production.
Defects can be comparatively subtle and typically exhibit less variance then the embossing on the items.
Furthermore, the links and the background both reflect the camera's flash.

%\input{optional/sec_3_1_old}

%\vspace{-5pt}
\section{Discovery of Limitations through Normalizing Flows}\label{sec:limits-of-nfs}
% JN: I think we need to say something about the fact that CS-flow wasn't the only SOTA and as well regarding GLASS there exists other if not as good, but still  good methods!!
When our project started in early 2023, one of the state of the art methods in unsupervised anomaly detection and -localization was a method called ``\emph{Fully Convolutional Cross-Scale-Flows for Image-based Defect Detection}'', here and after called ``CS-Flow''~\cite{RudolphWRW22_CSFLOW}.
Over the course of approx. 1.5 years, we would attempt to facilitate it to solve our own problem (see Section~\ref{sec:prob-and-dataset}).
In mid-2024, CS-Flow was superseded by a new state of the art method called ``\emph{A Unified Anomaly Synthesis Strategy with Gradient Ascent for Industrial Anomaly Detection and Localization}'', or, as its authors abbreviate it, ``GLASS''~\cite{ChenLLZ24_GLASS}.
As of writing this, GLASS is still undefeated.

In this section, we outline our journey of attempting to facilitate CS-Flow and adapting it to our problem.
This journey is characterized by an iterative and hypothesis-/evidence-based approach.
Our endeavors allowed us to unveil the intricate and inconspicuous issues related to data quality, choice of model, and evaluation approach.
For full disclosure, comprehensibility, and reproducibility, each step is associated with a notebook in the replication package~\cite{replication_package}.
The remainder of this section is dedicated to the larger steps in this process.
All experiments were run on an NVIDIA 8xH100/80GB machine.

% SH: Note to myself: Before I forget it, but as part of our contributions we should also outline lessons learned specifically with NFs (see my BigData presentation). I think that some of these could be very valuable, esp. if you really want/have to use an NF.

\subsection{Previous {SotA}: Convolutional Cross-Scale Normalizing Flow}\label{sec:csflow}
% SH: @Jonas: I think this is fine, but could be even more technically detailed (like choice of base distribution, how scores and anomaly maps are computed, ...). For an example, look at how deep I went with the next subsection which is for GLASS. The goal should be to have a similar depth, while trying to keep it short.
CS-Flow \cite{RudolphWRW22_CSFLOW} is a technique that processes image embeddings through a specialized normalizing flow to compute likelihoods of feature maps, which are then used to distinguish anomalies from normal samples.

The method begins by resizing each image into three different scales and passing them through a frozen feature extractor, EfficientNet-B5~\cite{TanL19_EffnetB5}, to generate three corresponding feature maps representing the image at multiple resolutions.
A key component of CS-Flow is a unique coupling block that captures interactions between feature maps at different scales.
These coupling blocks are stacked to form a normalizing flow, whose output is evaluated against a multivariate standard normal distribution with diagonal covariance matrix.
The resulting likelihood determines whether an image is anomalous, based on a threshold $\tau$.

In the paper, CS-Flow is evaluated on the MVTec AD dataset, achieving an average AUROC of $98.7$.
It reports state-of-the-art performance in eleven out of $15$ categories.
Furthermore, their approach processes feature maps in a fully convolutional (equivariant) manner, thus retaining positional information, enabling the identification of specific regions of anomalies.
To achieve this, an anomaly score is assigned to each local position $(i, j)$ of the feature map by aggregating values along the channel dimension using the  $L_2$-norm.
Regions with high norm values in the output feature tensors indicate potential anomalies.

\subsection{Current {SotA}: Global and Local Anomaly co-Synthesis Strategy}\label{sec:glass}
GLASS is a three-way discriminator (\emph{i.e.}, not a probabilistic model) that draws from all three branches of AD: embedding, reconstruction, and especially synthesis~\cite{ChenLLZ24_GLASS}.
It uses a frozen feature extractor, as well as a feature adaptor, which is a fully-connected network that learns a useful representation of those features.
It requires foreground masks of the objects of interest in order to synthesize so-called local anomalies.
For a local anomaly, GLASS blends the foreground mask with a randomly generated mask first.
Then, using a texture from the Describable Textures Dataset~\cite{CimpoiMKMV14_DTD}, the resulting patches are overlayed, thereby simulating an anomaly.
This results in pixel-accurate augmentations, making the entire method self-supervised.
Another type of synthetic anomaly that is generated are so-called global anomalies.
For each batch, GLASS takes the derivative of itself with respect to the nominal sample.
Then, Gaussian noise is added, and the sample is changed using truncated projection of the gradient (\emph{i.e.}, the sample is changed such that its loss under the current model increases).
Finally, the loss is computed along three branches:
The discriminator shall predict all zeros for nominal samples, all ones for global anomalies (\emph{i.e.}, anomalous everywhere), and all zeros except in the augmented regions for local anomalies.

Note that GLASS operates optionally under one of two distribution hypotheses:
the hypersphere and manifold hypotheses.
Under the first hypothesis, it is assumed that all nominal samples can be encompassed by a compact hypersphere and that all out-of-distribution samples have a distance from the center that is greater than that of any in-distribution sample~\cite{GoyalRJS020_DROCC}.
Under the second hypothesis, it is assumed that all nominal samples lie within a lower-dimensional, locally linear manifold distribution.
The local linearity and the fact that the manifold is homeomorphic to Euclidean space allows GLASS to define a distance under which a sample is considered to lie outside the in-distribution manifold~\cite{RuffGDSVBMK18_deep1clz}.
By synthesizing global anomalies, a nominal sample is gradually worsened until it is considered to be located outside the in-distribution manifold.
Interestingly, our dataset of coupling links conforms to a \emph{cyclic manifold}, because the objects themselves repeatedly assume previously-assumed angles~\cite{PlessS09_distr_hyp}.
This might also explain why we generally achieve better results by operating GLASS under the manifold hypothesis.

\begin{figure}
    \center % ska bytas till bilder med bakgrund sedan
    \includegraphics[width=0.475\textwidth]{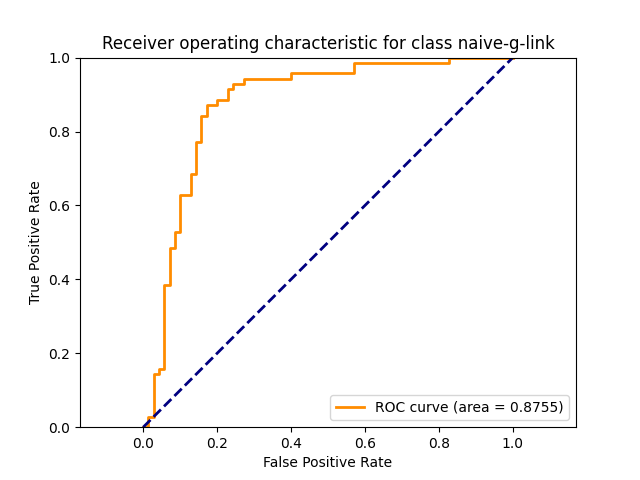}
    \quad
    \includegraphics[width=0.475\textwidth]{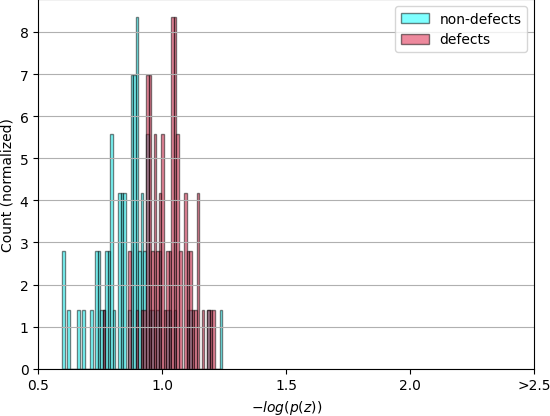}
    \caption{\label{fig:csflow-naive} ROC-curve (left) and anomaly scores (right) for the naive G-Link application under CS-Flow. Scores of nominal and anomalous samples overlap significantly.}
\end{figure}

\begin{figure}
    \center
    \includegraphics[width=1\textwidth]{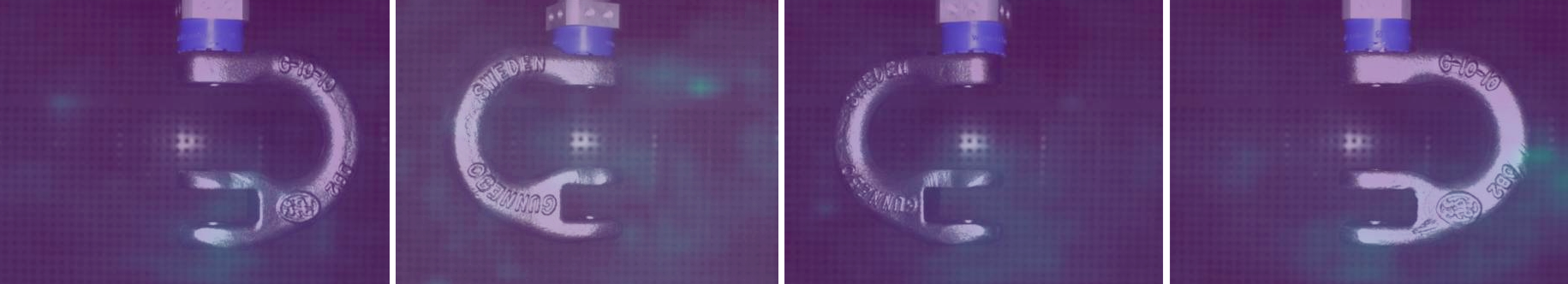}
    \caption{\label{fig:csflow-naive-maps} Two defective samples (left) and two nominal samples (right). Anomaly scores of nominal samples sometimes greatly exceed those of actual defective samples (increased gamma and contrast). Often, high scores are assigned to the background.}
\end{figure}

%\vspace{-10pt}
\subsection{Na\"ive Application to Raw Dataset}
The first step was to attempt to use CS-Flow as an off-the-shelf solution and apply it to raw imagery of our dataset of forged coupling links.%
%\footnote{Our approach can be replicated using the notebook \texttt{cs-flow\_naive-g-link}.}
We have not made changes to the default configuration, which was used to achieve state-of-the-art results on the MVTec AD dataset.
By default, images are resized to 768x768 pixels.
Two smaller versions of the same image are used, one in 384x384 and one in 192x192, as CS-Flow facilitates a multi-scale approach that extracts and combines features from all three images per sample.
The default architecture comprises four coupling blocks, applies gradient-clipping, and uses 1,024 units in the fully-connected networks that are the conditioners to each coupling block.
The training runs for 240 epochs by default, with an intermediate evaluation step every 60 epochs.

\begin{table}
\caption{\label{tab:results}Summary of AUROC scores across models and datasets. Note that the AUROC of the MVTec AD datasets are based on a rotated dataset as described in Section \ref{ssec:controlled-rotation}.}

  \centering
  \begin{tabular}{p{1.5cm} p{2.85cm} p{1.4cm}}
    \toprule
    Model & Trained On & AUROC \\
    \midrule
    \multirow{5}{=}{CS-Flow} & MVTec AD & $0.631$ \\
                             & MVTec AD (rot.) & $0.939$ \\
                             & Na\"ive G-Link & $0.871$ \\
                             & Masked G-Link & $0.802$ \\
                             & CE G-Link & $0.867$ \\
    \midrule
    GLASS & CE G-Link & $\mathbf{0.941}$ \\
    \bottomrule
  \end{tabular}
  
\end{table}

The evaluation of the obtained results from this step promised a skillful model.
Typically, the reported AUROC on the test set would be $\geq0.87$ and the plot of the anomaly scores (which are derived from the negative log-likelihood an image produces) would indicate a somewhat good separability of scores.
An example histogram and ROC-curve are shown in Figure~\ref{fig:csflow-naive} and results in Table~\ref{tab:results}.
%Recall that AUROC is the probability that a randomly selected positive instance is scored higher than a randomly selected negative instance by the trained model.
%Upon inspecting the evaluation routines of CS-Flow, we noticed that the optimal threshold for separating classes by log-likelihood is computed on the same set it is applied to. %, which is problematic.
%While this is common practice across many publications (e.g.,~\cite{RudolphWRW22_CSFLOW,MaLHWYMM23_SANF_ad,ZhouXSSS25_MSFlow}), it lacks expressive power with regard to a potential classifier (see Section~\ref{ssec:robust} for a robust estimation of factory-level performance).
% One way to mitigate this is to have a 3rd holdout set: train on first set, evaluate (and determine optimal threshold) on second, then report accuracy on 3rd set.

\noindent
The next evaluation step was to inspect the anomalous regions.
Four typical examples are shown in Figure~\ref{fig:csflow-naive-maps}.
It becomes apparent that most of the low-likelihood pixels are assigned to the background.
The highlighted anomalous regions are not concordant with the actual location of the defect.
Even worse, it would appear that the residual anomaly maps roughly cover the area where the coupling link is \emph{usually} spatially present.
What we mean by that is that most images in our dataset show the coupling link rotated (to a varying degree) along the fixture's axis to one or the other side.
One could argue that if we were to interpolate all images of our dataset, the result would perhaps show the silhouette of a somewhat ``double-link'', which represents the average of it.
Since a normalizing flow learns a complex probability distribution, it would approximate this average image, which is in disparity with any single image.
This in turn allows us to conclude and hypothesize three things:
\begin{enumerate}
    \item We hypothesize that strong object-variance can have detrimental effects when the underlying model is a deep density estimator.
    \item The background should be blacked out in order to prohibit the model from assigning high (or any positive) anomaly score to it.
    \item We have a Husky--Snow problem, likely caused by too-large spatial object variance.
\end{enumerate}

\noindent
The ``Husky--Snow'' problem, sometimes also referred to as ``Clever--Hans`` problem, is used to demonstrate how a model may learn unintended background features\,---\,in this case, a model distinguishing between wolves and huskies using background snow presence\,---\,rather than the animal itself (or discriminative features thereof) as the distinguishing factor.
More formally, these so-called local interpretable model-agnostic explanations were introduced by Ribeiro et al.~\cite{Ribeiro0G16_huskySnow}.
What this means for us is that the model has not learned any useful representation of the underlying problem.
It might have memorized specialized patterns or shortcuts from the training set without extracting broader features to achieve the relatively high AUROC.
One may also consider the average image (silhouette) as intrinsic noise in the training set: Without sufficient capacity for generalization (or compensating invariances) in the model, it appears we are posing a contradictory problem in the first place.
Recall that in a standard normalizing flow architecture, the available learning capacity is restricted to the conditioners, which learn the parameters $\theta$, required by the transformation $T(\mathbf{x};\theta)$.
In other words, between the (frozen) feature extractor and the flow itself, CS-Flow does not currently offer some form of learnable embedding.

% SH: UPDATE: I incorporated this into the previous paragraph, but keep here for reference! Also, I think we're good on the visualizations here!
%\subsection{Husky--Snow problem}
% SH: @Jonas: This should probably be merged with previous subsection! 
% SH: @Jonas, feel free to fill this out with some general terms, I will fill in the specifics (how it appeared in our case).
% SH: @Jonas: Note that this is sometimes also called "Clever-Hans" or "Clever-Horse" problem
% - Presented results (image AUROC ~0.85 perceived as positive initially).
% - Visualization clearly demonstrating Husky-Snow issue: anomalous regions on background.
% - Show statistical overlap of distributions of likelihood scores.

\subsection{Segmentation and Background Removal}
% SH: I will write this:
% - Motivation/intention clearly described (SAM-based segmentation, mixed model approach).
% - Surprising negative impact: decreased AUROC (0.8), more overlapping distributions.
% - Reasoned analysis: Normalizing flow "prefers" commonplace positions due to learned spatial distributions.
%
The primary consequence from our na\"ive application to the coupling links dataset is to remove (or to black out) the background from each image, including removal of the sometimes visible fixture atop.
However, since we find ourselves in an inherent unsupervised setting, manual removal of the backgrounds for each image would be too costly.
While not technically completely unsupervised, we used GroundedSAM~\cite{Ren2024_GroundedSAM} to segment the bulk ($\approx\!98$\%) of our dataset.
GroundedSAM is an open-vocabulary detection and segmentation model.
It facilitates textual prompts to instruct SAM (Segment Anything,~\cite{KirillovMRMRGXW23_SAM}) to detect and segment corresponding objects.
While it took a handful of different prompts, this final prompt ``\emph{the curved metal part in the center of the image without the blue ring}'' was used to segment the entire dataset.
As a micro ablation, we segment images from the MPDD dataset~\cite{JezekJBDS21_MPDD} using similarly trivial prompts and on-par segmentation results%
\footnote{Among our contributions are two command-line applications that enable efficient mass-wise processing of images and their masks using a single prompt.}
.
% SH: TODO: If there is time, we should actually do this and include!

\begin{figure}[H]
    \center
    \includegraphics[width=0.99\textwidth]{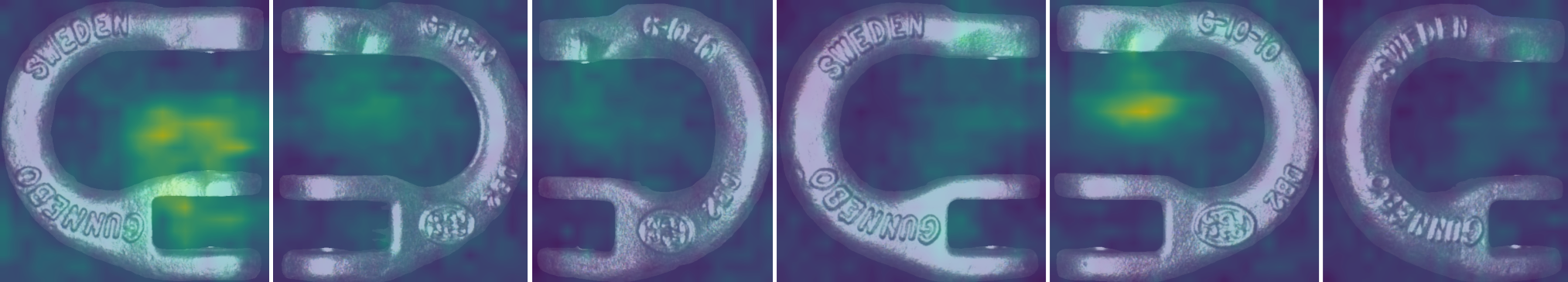}
    \caption{\label{fig:csflow-masked-maps} Examples of spurious anomaly maps of defect samples using the masked dataset of coupling links (increased gamma and contrast).}
\end{figure}

\noindent
Although the performance of training a vanilla CS-Flow on the segmented data has declined, the typical AUROC scores in the range of $[0.79,0.81]$ would still suggest a somewhat skillful model.
Upon inspecting the identified anomalous regions, we realize that higher scores are still assigned to the now-black background in many cases, meaning a worse version of the Husky--Snow problem is still present.
Since the model cannot learn any shortcuts through spurious features of the background any longer, this result explains the lower AUROC and presents evidence for our silhouette theory.
In short, it appears that the varying angles of the object seem to be disadvantageous for the chosen model.
%
% SH: The following is optional:
%While impractical for our dataset, one could apply techniques for image aligning and -warping, such as the scale-invariant feature transform~\cite{Lowe99_SIFT} in order to produce more homogeneous imagery and reduce the dataset's natural variance.
%Robustness may also alternatively introduced by a scale- and location-invariant feature extractor (hard) or a non-linear decision rule.

\begin{figure}
    \center
    % SH: This figure is too wide by design, so that the colorbar sticks out
    \includegraphics[width=1.075\textwidth]{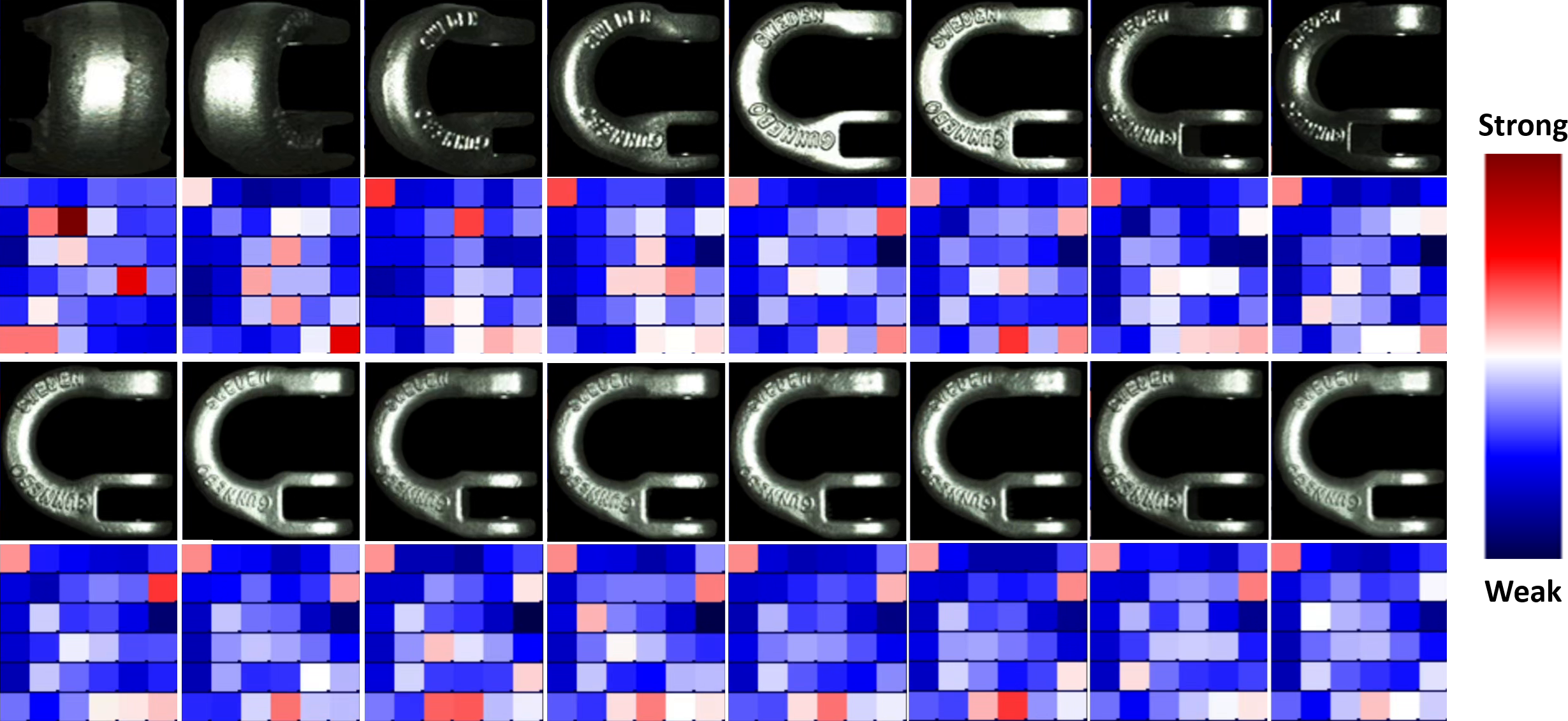}
    \caption{\label{fig:equivariance} A highly varying dataset causes an equivariantly varying feature map, too. Aggregated were the 6x6x304 feature maps, each using $\max()$. The lowest low is the supremum of the maximum activations across all patches and across all images (\emph{i.e.}, $\max()$ is the global maximum).}
\end{figure}

\subsection{Controlled Rotation Experiment on MVTec AD}\label{ssec:controlled-rotation}
%
% SH: Sub-section is good for now!
%
Following our na\"ive approach, we hypothesized that a pronounced lack of object invariance leads to detrimental performance.
However, it remains unclear whether the same holds true for image invariance—particularly relevant for equivariant models.
Convolutions inherently provide some approximate local translation invariance.
Pooling further amplifies this, capturing only the maximum, average, or minimum in local regions.
Thus, for limited translation and scale variance, equivariant models offer partial image invariance. 
To quantify translation invariance, we designed a two-fold controlled experiment using a vanilla CS-Flow model with the ``cable'' category from the MVTec AD dataset~\cite{BergmannBFSS21_MVTEC_AD}.
We first trained on unmodified nominal data containing three strands: a central yellow strand at the top and blue and brown strands in the left and right corners.
For testing, we added rotated versions (90°, 180°, 270°) of these nominal images.
In a second test, training images were rotated using the same angles, while inference used original, as well as non-rotated images.
One goal of this test was to find out to what degree CS-Flow is resilient to rotation variances.
Ideally, a rotated nominal sample should not yield anomaly scores higher than typical nominal test images, indicating robust invariance.
For the first experiment, the optimal threshold was $\approx\!0.957$ with AUROC $\approx\!0.631$ across all anomaly categories (e.g., bent wire, poked insulation).
Rotated images had anomaly scores between $\approx\!1.28$ and $\approx\!1.64$, resulting in all of the rotated images being wrongly classified as anomalous.
In contrast, the second experiment achieved a significantly higher AUROC ($\approx\!0.936$) and an optimal threshold of $\approx\!1.251$.
These two tests indicate that there is only very low robustness with regard to rotational invariance.
Only one out of $58$ non-rotated nominal samples was incorrectly classified (score $\approx\!1.278$).
Thus, the second model generalized better, successfully handling rotated images. 
In summary, we demonstrated that rotation-based image invariances can be effectively handled when the training set explicitly includes such variance.
We hypothesize that careful and limited artificially introduced image variance helps equivariant models generalize better.
However, this approach may simultaneously increase model fragility through learning contradictory information—especially prominent when object invariance is introduced. 
Figure~\ref{fig:equivariance} visualizes activation magnitudes of latent features under an equivariant feature extractor (EfficientNet-B5,~\cite{TanL19_EffnetB5}).
The first row shows rotated coupling links, revealing activations tightly localized around object regions.
The second row depicts coupling links with only minimal viewpoint changes.
While activation intensities vary slightly, the activation locations remain consistent as the object's position changes minimally.

\subsection{Restriction to Similar Angles}
%
% This is currently computing. I will wait for the results and see how (if) this fits into the paper.
%
Since CS-Flow is a deliberate equivariant pipeline, we alter our dataset one more time, by limiting the imagery to those shot from a very similar, front-facing angle%
%\footnote{The results are computed in \texttt{cs-flow\_one-angle-g-link}.}
.
The idea is, again, to reduce the natural variance in the data.
However, that same restriction also applies to the test data, meaning that we alter and simplify our initial problem.
Therefore, the results cannot (and should not) be compared to the others.
However, this method achieves an AUROC of $\approx\!0.909$.
The qualitative evaluation shows that the anomaly maps often are closely located to the true region of the anomaly.
However, having occasional spurious high anomaly scores assigned to the background is still a problem in this model.

% SH: I think we're there!
%\subsection{Limits of Equivariant Features and Normalizing Flows Capacity}
% SH: TODO: Conclude previous subsection and merge with this one, once replication finished
% SH: Here we hypothesize:
% - Present clear visualizations: Activations of EfficientNet-B5 spatially shift with object rotation. (I have a GIF that we can decompose into some images)
% - 

\section{Robust Dataset Preparation and Performance Estimation}\label{sec:robust-estimation}
We produce a segmented, centered, and padded version of our original dataset (``center-embedded'' or ``CE'' short).
In it, all objects are segmented and cropped to the minimum bounding box first.
Then, we determine the minimum width and height that can encompass all coupling links, regardless of their spatial arrangement.
We find that a resolution of $600$x$800$ suffices.
Finally, padding is added to each image such that the object remains vertically and horizontally centered.
This way, the coupling links do not rotate longer around the fixture's axis, but rather around the image's center.
This allows us to effectively reduce natural variation while retaining all information.

\subsection{Results using CS-Flow}
We perform a last test using CS-Flow and this dataset, as previously, images were resized by it to a fixed resolution, which likely introduced more unwanted object variance, as narrow-appearing objects are perhaps stretched very wide%
%\footnote{The corresponding notebook is called \texttt{cs-flow\_ce-g-link}.}
.
With an AUROC of $\approx\!0.867$, the model does surprisingly not perform better than our na\"ive tests.
Spurious anomalous regions pose still a problem.
It would finally appear that the too-large natural object variance makes the flow model inherently fragile as it attempts to learn contradictory information depending on object position.
Recall that vanilla Normalizing Flows, even with equivariantly extracted features, lack a distinct embedding space robust enough to cope with large object-level variations like rotation and position shifts.

\subsection{Results using GLASS}
\begin{figure}
    \centering
    \includegraphics[width=1.0\textwidth]{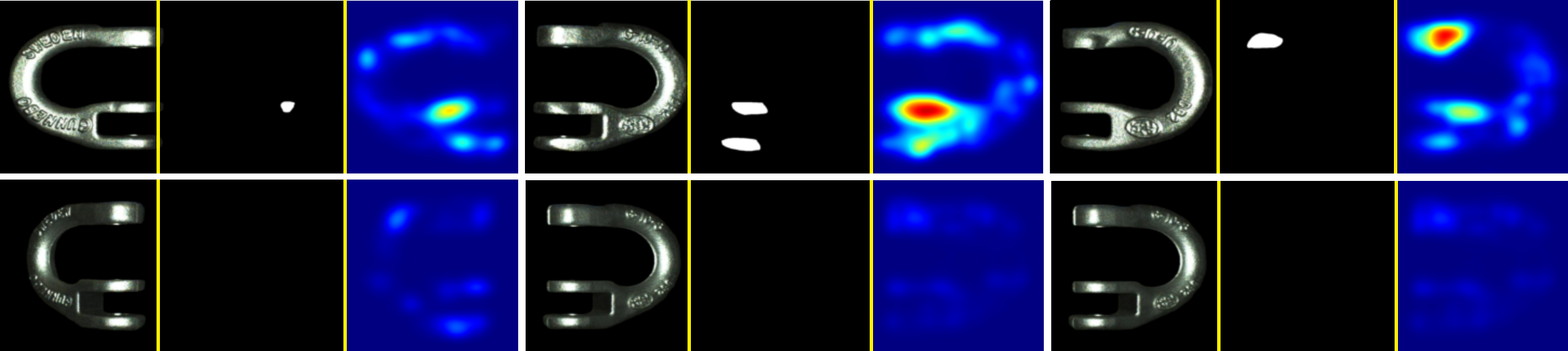}
    \caption{Three anomalous (top row) and three nominal samples (bottom row) as predicted by GLASS. The ground truth mask for each is in the middle.}
    \label{fig:glass}
\end{figure}

%
% SH: TODO
Contrary to CS-Flow, GLASS is not a multi-scale model.
Instead, it accepts images of a single resolution only.
We identified a champion model that uses a lowered resolution of $160$x$160$ that achieves an image-level AUROC of $\approx\!0.941$.
The qualitative inspection of the results shows strong agreement between true and predicted anomaly location.
Similarly, anomaly scores are practically zero for nominal parts (see Figure~\ref{fig:glass}).
While sometimes the predicted anomaly scores are too high for some regions, GLASS never assigns any scores to the background, so it seems the silhouette-problem vanished.
Also, judging by the qualitative results, it appears that the strong object-variance in our dataset is not problematic here, either.
Interestingly though, GLASS also uses an equivariant (frozen) feature extractor, here WideResNet-50 by default.
However, compared to Normalizing Flows, GLASS is a purpose-built model for AD, with two nonlinear high-capacity components, a feature adaptor (learning a suitable embedding and salient features), as well as a dense discriminator.
We conclude that it is this architecture that allows the model to learn a more generalized representation of our problem.

%\vspace{-5pt}
\subsection{Obtaining Unbiased Robust Performance Estimates}\label{ssec:robust}
Once we are confident in our model and want to take it from the laboratory to the factory, we require robust estimates as to the usefulness for unsupervised industrial inspection.
So far, we have only computed the AUROC.
However, its usefulness lies in comparing models and/or datasets and is limited to the dataset it was determined on.
In order to estimate other metrics, such as (balanced) accuracy, F1, precision and recall, etc., we have to consider the previously computed optimal threshold and use it in a decision rule on a \emph{new} dataset.
The most common choice for selecting a threshold is to use Youden's index to determine the maximum value of a ROC curve. %~\cite{Youden1950}.
However, common alternatives are to select the threshold by, e.g., maximizing the sensitivity/specificity trade-off, maximizing a particular cost-based metric, or another criterion, such as taking into account the real-world cost of a false positive/negative (e.g., ``\emph{what is the cost of an overlooked defective part?}'').

\begin{figure}
   \centering
   \includegraphics[width=\textwidth]{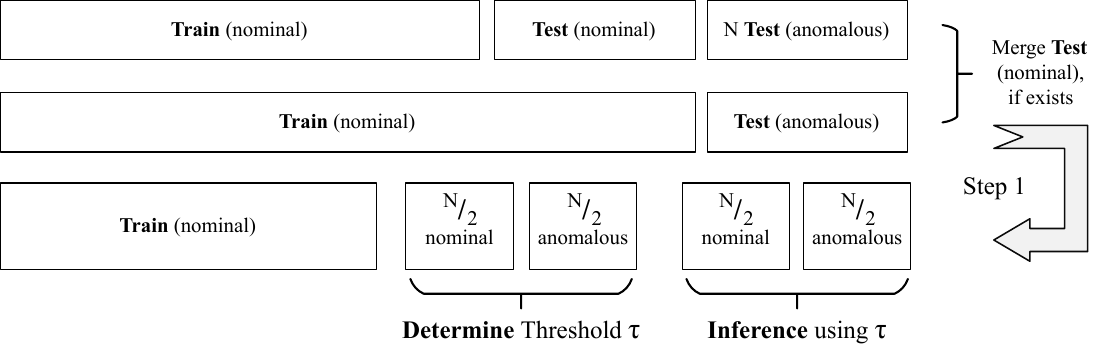}
   \caption{\label{fig:steps_1_2} Repartitioning of the training and testing data is the first step in the pipeline for obtaining robust factory-level estimates.}
\end{figure}

\noindent
In the following, we describe the necessary steps that allow obtaining an unbiased estimator for the empirical risk. %, the first of which is depicted in
%(Figure~\ref{fig:steps_1_2}).
These steps are in accordance with existing guidelines that are recommended also in cases of limited samples sizes~\cite{vabalas2019}.
We assume that a three-part dataset exists to begin with: nominal training samples, as well nominal and anomalous test samples.
We further assume that either the nominal test samples were drawn i.i.d. from the same distribution as the train samples, or that no nominal test set exists (second row, Figure~\ref{fig:steps_1_2}).

\begin{enumerate}[font=\bfseries,leftmargin=40pt]
    \item[Step 0)] (Optional) If a nominal test set exists and the i.i.d. assumption holds, merge it into the nominal train set.
    \item[Step 1)] Perform a random, but deterministic three-way split. Separate the anomalous test data randomly into one half each for determining the threshold and for inferencing. Retain two partitions of the same size of nominal samples. Separate the remaining nominal samples into the training data.\footnote{If the image set contains multiple recordings of the same object, as is the case in our ``G-link’’ dataset, one must ensure images of the same object are not in both train and test to prevent information leakage.}
    \item[Step 2)] Train model on remaining nominal data until convergence.
    \item[Step 3)] Predict anomaly scores on \emph{threshold} dataset and determine the optimal threshold $\tau$ for separating nominal/anomalous samples.
    \item[Step 4)] Apply threshold $\tau$ to \emph{inference} dataset and compute desired metrics. Record all results.
    \item[Step 5)] Repeat $K$ times for $K$-fold cross-validation (go to Step 1).
\end{enumerate}

\noindent
As for the number of repeats, a typical recommendation is at least $5$--$10$.
However, there exist insightful empirical guidelines for cross-validation experimentation for obtaining stable model assessments~\cite{Bouckaert04_KfoldCV}.
Choosing partition sizes different from $\sfrac{N}{2}$ in Step 1 will not result in an unbiased estimation.
This is because ROC curve and AUROC can be overly optimistic for severely imbalanced classification problems and unsuitable when false negatives and false positives have significantly different costs~\cite{branco_predmodelsurvey}.

\section{Discussion and Future Work}\label{sec:discuss-fw}
Approaching the G-link problem na\"ively and without prior assumptions inspired us to look beyond too-good-looking AUROCs and to iteratively identify and correct a problem that was inherently rooted in data quality and model choice.
We identify an overall fragility of likelihood-based methods without representation learning (embedding) under real industrial conditions.
We learn that the amount of natural variance present in a dataset strongly affects model performance and its ability to generalize and discriminate.
Lastly, we highlight the importance of inspecting reported metrics and examining their applicability to factory-level operations.

% Clearly discuss and highlight practical and theoretical implications:
%
% - Overall fragility of normalizing flows for real industrial conditions.
% - Importance of learned embedding (beyond just CNN features extraction) robust to object-view invariance.
% - Importance of explicit anomaly synthesis (local and global).
% - Initial limitations of GLASS-based global anomaly synthesis (gradient ascent) and the promising idea of controlled global anomaly synthesis leveraging the known manifold topology of normalizing flows.
%
% Discuss briefly the preliminary early-stage experiments about explicitly integrating flows to control global anomaly synthesis—highlighting your early findings as promising future work avenue.

% \section{Future Work}
%
We hypothesize that the synthesis of local and global anomalies is what allows GLASS to better concretize its in-distribution manifold.
While the synthesis of local anomalies was done many times in previous works, the global anomaly synthesis strategy is novel.
However, GLASS is specialized to imagery and, as such, the procedure using clamped gradient ascent is hard to control (hence the introduction of stochasticity).
We have since begun to work on a flow-based model that exploits the fact that under a known base distribution (e.g., isotropic Gaussian), we can take controlled steps that allow us to gain insights into the geometry of the latent data manifold.
This technique, in combination with controlled noise, allows us to effectively traverse the loss space linearly along all dimensions and to controllably synthesize anomalies.

% SH: There are 2 parts here, I will at least write the 2nd.
%
% 1st part:
% Recap contributions succinctly: 
%
% - Highlight insight of NF limitations.
% - Highlight concrete GLASS improvement (practical usability for real-world industrial data not lab-perfect ones).
% - Clearly suggest immediate implications and applicability for practitioners.
%
%
% 2nd part:
% Clearly describe future work: 
%
% - Rigorous exploration of controlled global anomaly synthesis using normalizing flows to further enhance robustness and manifold understanding.
% - Extending methodology to other difficult industrial inspection domains.

\begin{credits}
\subsubsection{\ackname} %SH: Here we need to mention Gimic, the project, its funding, etc., as well people that helped with anything \emph{but} writing of this paper.
%\emph{The acknowledgments have been anonymized and are currently withheld in order not to reveal the authors' identities.}
Our project ``\emph{In-line visual inspection using unsupervised learning}'' is a 2022 Vinnova project for Advanced Digitization, application number 2022-03018.
We would like to sincerely thank our project partners and co-financers, namely Linnaeus University and its High-Performance Computing Center, Gimic AB, SKF, and Gunnebo Industries.

\subsubsection{\discintname}
The authors have no competing interests to declare that are relevant to the content of this article.
%It is now necessary to declare any competing interests or to specifically state that the authors have no competing interests. Please place the statement with a bold run-in heading in small font size beneath the (optional) acknowledgments, for example:
%The authors have no competing interests to declare that are relevant to the content of this article. Or: Author A has received research grants from Company W. Author B has received a speaker honorarium from Company X and owns stock in Company Y. Author C is a member of committee Z.
\end{credits}

\bibliographystyle{splncs04}

\end{document}